\documentclass[sigconf, screen]{acmart}

\usepackage{caption,tabularx,booktabs, arydshln}
\usepackage{xcolor}
\usepackage{comment}
\usepackage{multirow}
\usepackage{adjustbox}
\AtBeginDocument{%
  \providecommand\BibTeX{{%
    \normalfont B\kern-0.5em{\scshape i\kern-0.25em b}\kern-0.8em\TeX}}}

\begin{document}

\title{REACT 2025: the Third Multiple Appropriate Facial Reaction Generation Challenge}

\author{Siyang Song}
\affiliation{%
  \institution{University of Exeter}
\country{United Kingdom}
  }
  \email{s.song@exeter.ac.uk}

\author{Micol Spitale}
\affiliation{%
  \institution{Politecnico di Milano}
  \city{Milan}
  \country{Italy}
  }
  \email{micol.spitale@polimi.it}

\author{Xiangyu Kong}
\affiliation{%
  \institution{University of Exeter}
  \country{United Kingdom}
  }
  \email{xk219@exeter.ac.uk}

\author{Hengde Zhu}
\affiliation{%
  \institution{University of Leicester}
  \country{United Kingdom}
  }
  \email{hz204@le.ac.uk}

\author{Cheng Luo}
\affiliation{%
  \institution{King Abdullah University of Science and Technology}
  \country{Saudi Arabia}
  }
  \email{cheng.luo@kaust.edu.sa}

  \author{Cristina Palmero}
\affiliation{%
  \institution{King's College London}
  \city{London}
  \country{United Kingdom}}
\email{cristina.palmero@kcl.ac.uk}

  \author{German Barquero}
\affiliation{%
  \institution{Universitat de Barcelona}
  \city{Barcelona}
  \country{Spain}}
\email{germanbarquero@ub.edu}

  \author{Sergio Escalera}
\affiliation{%
  \institution{Universitat de Barcelona}
  \city{Barcelona}
  \country{Spain}}
\email{sergio@maia.ub.es}

  \author{Michel Valstar}
\affiliation{%
  \institution{University of Nottingham}
  \city{Nottingham}
  \country{United Kingdom}}
\email{michel.valstar1@nottingham.ac.uk}

  \author{Mohamed Daoudi }
\affiliation{%
  \institution{IMT Nord Europe}
  \city{Villeneuve d'Ascq}
  \country{France}}
\email{mohamed.daoudi@imt-nord-europe.fr}

  \author{Tobias Baur}
\affiliation{%
  \institution{University of Augsburg}
  \city{Augsburg}
  \country{Germany}}
\email{tobias.baur@uni-a.de}

  \author{Fabien Ringeval}
\affiliation{%
  \institution{Université Grenoble Alpes}
  \city{Grenoble}
  \country{France}}
\email{fabien.ringeval@imag.fr}

\author{Andrew Howes}
\affiliation{%
  \institution{University of Exeter}
  \city{Exeter}
  \country{United Kingdom}}
\email{andrew.howes@exeter.ac.uk}

\author{Elisabeth Andrè}
\affiliation{%
  \institution{University of Augsburg}
  \city{Augsburg}
  \country{Germany}}
\email{andre@uni-a.de}

\author{Hatice Gunes}
\affiliation{%
  \institution{University of Cambridge}
  \city{Cambridge}
  \country{United Kingdom}}
\email{hatice.gunes@cl.cam.ac.uk}

\renewcommand{\shortauthors}{Song et al.}

\begin{abstract}

In dyadic interactions, a broad spectrum of human facial reactions might be \textit{appropriate} for responding to each human speaker behaviour. Following the successful organisation of the REACT 2023 and REACT 2024 challenges, we are proposing the REACT 2025 challenge encouraging the development and benchmarking of Machine Learning (ML) models that can be used to generate multiple \textit{appropriate}, \textbf{diverse}, \textbf{realistic} and \textbf{synchronised} human-style facial reactions expressed by human listeners in response to an input stimulus (i.e., audio-visual behaviours expressed by their corresponding speakers). As a key of the challenge, we provide challenge participants with the first natural and large-scale multi-modal MAFRG dataset (called MARS) recording 137 human-human dyadic interactions containing a total of 2856 interaction sessions covering five different topics. In addition, this paper also presents the challenge guidelines and the performance of our baselines on the two proposed sub-challenges: Offline MAFRG and Online MAFRG, respectively. The challenge baseline code is publicly available at \url{https://github.com/reactmultimodalchallenge/baseline_react2025}.

\end{abstract}

\maketitle


\section{INTRODUCTION}


Recent advances in generative models have boosted numerous automatic human behaviour synthesis systems, including dialogue systems (e.g., Large Language Models (LLMs)) \cite{radford2018improving,liu2024deepseek}, audio/text-driven talking face solutions \cite{zhang2021facial,deng2025degstalk} and human body motion generation \cite{sun2025beyond}. While facial reactions play a key role for human in conveying their intentions and emotions in human-human interactions, they are characterized by the complex and variable nature of interpersonal communications \cite{hess1998facial}. Due to lack of comprehensive benchmark and dataset, automatic human facial reaction generation (FRG) is still an underexplored research area, despite that the ability to generate realistic and contextually appropriate human-style facial reactions is essential for developing advanced humanoid virtual agents/robots \cite{wang2021examining,savva2019habitat}. Early FRG approaches \cite{nojava2018interactive,huang2018generative,huang2018photorealistic,song2022learning,shao2021personality,zhou2022responsive,ng2022learning} initially posited that each behaviour (e.g., facial behaviour or audio-visual behaviour) expressed by a speaker (called speaker behaviour) could only trigger human listeners to express a single facial reaction, assuming FRG as a `one-to-one mapping' problem. These efforts aim to accurately replicate the ground truth (GT) listener facial reactions, such as facial reaction sketches \cite{huang2017dyadgan}, photorealistic expressions \cite{huang2018photorealistic}, facial landmark sequences \cite{song2022learning,shao2021personality}, or full face videos \cite{Ng_2022_CVPR}, derived from the observed facial or audio-visual behaviours of the speaker.

In real-world human-human interactions, an identical speaker behaviour may elicit varied facial reactions expressed from different listeners or the same listener under different contexts \cite{mehrabian1974approach}, suggesting that FRG is fundamentally a `one-to-many mapping' task \cite{song2023multiple}. As a result, the recently introduced Multiple Appropriate Facial Reaction Generation (MAFRG) task hypothesizes that each speaker behaviour may trigger human listeners to express different but realistic and appropriate facial reactions (AFRs) depending on external and their internal contexts \cite{song2023multiple}. The organization of the REACT 2023 and REACT 2024 challenge \cite{song2023react2023,song2024react} facilitated the creation of several successful solutions \cite{xu2023reversible,luo2024reactface,zhu2024perfrdiff,hoque2023beamer,liang2023unifarn,yu2023leveraging,nguyen2024vector,liu2024one,zhu2025perreactor,nguyen2025latent,tran2024dim,hu2024robust,lv2025hierarchical,nguyen2024multiple,dam2024finite} for both online and offline MAFRG tasks. Specifically, they focus on addressing the ill-posed training problem (i.e., \textbf{one input} speaker behaviour may correspond to \textbf{multiple labels} (AFRs)) by learning either a distribution \cite{hoque2023beamer,liang2023unifarn,luo2024reactface,xu2023reversible,yu2023leveraging} or a codebook \cite{nguyen2024vector,liu2024one} to represent multiple plausible AFRs, from the input speaker behaviour, which reformulate the `one-to-many training' problem as the `one-to-one training' task.

However, both REACT 2023 and REACT 2024 challenges only include modified/segmented audio-visual clips originally collected by the NoXI \cite{cafaro2017noxi}, UDIVA \cite{palmero2021context} and RECOLA \cite{ringeval2013introducing} datasets, all of which were recorded for the purposes other than the MAFRG task. Consequently, their data and labels limit the development of more advanced MAFRG systems. Following the conclusion of both challenges, many researchers approached the organisers to request a natural and well-annotated multi-modal dataset specifically recorded for the MAFRG task. In response to this interest, we propose the third multiple appropriate facial reaction generation (MAFRG) challenge - REACT 2025, aiming to advance the current state-of-the-art MAFRG solutions using \textbf{the first specifically collected MAFRG dataset that contains natural human-human interaction scenarios and rigorous AFR labels.}

The REACT 2025 challenge follows previous REACT challenges to invite participants to attend offline MAFRG and online MAFRG tasks, both requiring to generate multiple AFRs in response to each given speaker audio-visual behaviour. Each generated AFR clip is represented by a face video and its corresponding multi-channel facial primitive time-series consisting of 25 frame-level facial attributes, i.e., 15 action units (AUs), 8 facial expressions, as well as valence and arousal intensities. It re-employs the metrics defined in \cite{song2023multiple} to evaluate four aspects of the submitted models in terms of their generated AFRs, namely: appropriateness, diversity, realism and synchrony. Participants are required to submit their developed model, checkpoints and well-explained source code, accompanied by a paper describing their proposed methodologies and the achieved results. Only contributions that meet the pre-determined requirements, terms and conditions \footnote{https://sites.google.com/view/react2025/home} are eligible for participation. The organisers do not engage in active participation themselves, but instead undertake a re-evaluation of the findings of the systems submitted to both sub-challenges. The ranking of the submitted models depend on two metrics:  correlation (FRCorr) of the generated facial reaction attributes and facial reaction realism (FRRea) of the generated facial reaction video clips, for both sub-challenges. In summary, the main contributions and novelties of this challenge are listed as follows: 
\begin{itemize}

    \item Introducing and sharing the novel and well-annotated REACT 2025 Multi-modal Challenge Dataset (called Multi-modal \textbf{M}ultiple \textbf{A}ppropriate \textbf{R}eaction in \textbf{S}ocial Dyads (MARS) Dataset), together with the recorded multi-modal audio, visual and newly introduced EEG features, as well as objectively-annotated ground-truth (GT) AFRs and self-reported personality traits of every given speaker behaviour.

    \item Presenting first open-source baseline models on MARS for generating multiple AFRs in response to each multi-modal speaker behaviour under various dyadic interaction scenarios, where the duration of speaker behaviours are not equal. Consequently, the REACT 2025 challenge requires the developed MAFRG models to be able to process variable-length inputs, aiming to simulate real-world human-human interactions where speakers may express the same intention with behaviours of different ways and rates, which is different from previous REACT challenges that only require to predict equal-length AFRs from equal-length speaker behaviours.

    
\end{itemize}


\begin{figure}[tb]
    \centering
    \includegraphics[width=1\columnwidth]{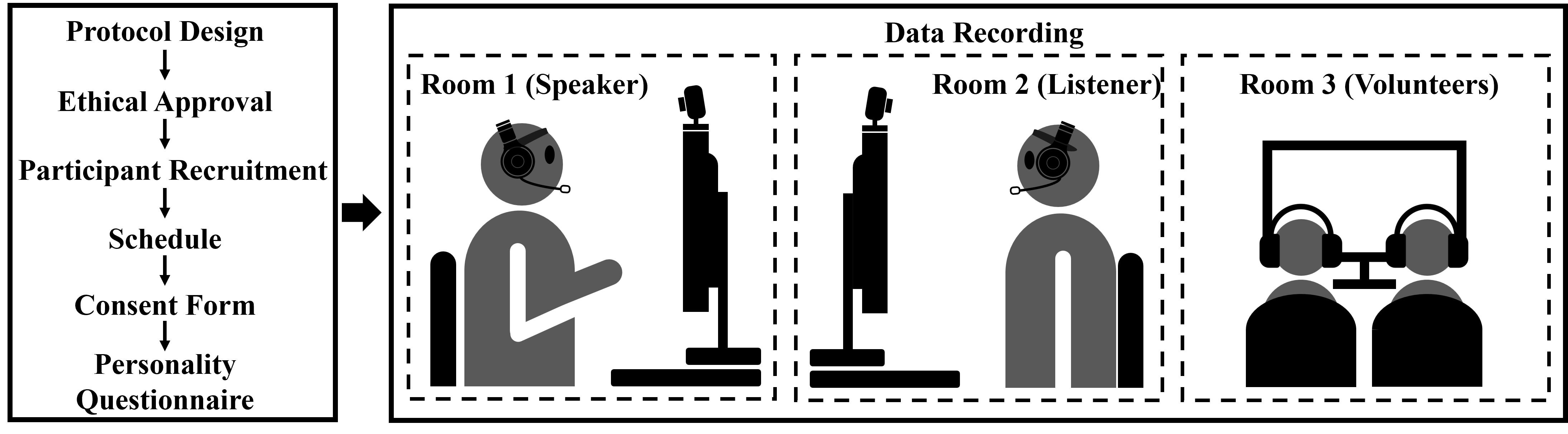}
    \caption{Illustration of the data collection scenario of the MARS dataset. The left side outlines the preparatory steps, including protocol design, ethical approval, scheduling, obtaining participant consent, and completing a personality questionnaire. The right side illustrates the physical data collection setup, where a pair of human speaker and listener sit in front of PCs to conduct a video conference in the context of several pre-defined interaction tasks.}
    \label{fig:diagram}
\end{figure}

\begin{table*}[h]
    \centering
    \caption{Demographics of participants in MARS dataset. $^{*}$: The highest levels of educational attainment are categorised as B: Below Undergraduate, U: Undergraduate Degree, M: Master's Degree, D: Doctoral Degree.}
    \begin{adjustbox}{width=2\columnwidth,center}
        \begin{tabular}{cccc c cccc c cccc c ccc}
            \toprule
             \multirow{2}*{Role} & \multicolumn{3}{c}{Gender} & & \multicolumn{4}{c}{Age Group} & & \multicolumn{4}{c}{Highest Degree$^{*}$} & &  \multicolumn{2}{c}{Mother Language} & \multirow{2}*{Total} \\
            \cline{2-4} \cline{6-9} \cline{11-14} \cline{16-17}
            & Male & Female & Others & & $\leq$20 & 21-30 & 31-40 & $\geq$ 41 & & B & U & M & D & & English & Others & \\
            \midrule
            Speaker & 10 & 13 & 0 & & 2 & 18 & 2 & 1 & & 2 & 4 & 14 & 2 & & 11 & 12 &  23 \\
            Listener & 71 & 65 & 1 & & 18 & 81 & 26 & 12 & & 26 & 29 & 58 & 24 & & 47 & 90 & 137 \\
            \bottomrule
        \end{tabular}
    \end{adjustbox}
    \label{tab:demographics}
\end{table*}

\begin{figure}[htb!]
    \centering
    \includegraphics[width=1\columnwidth]{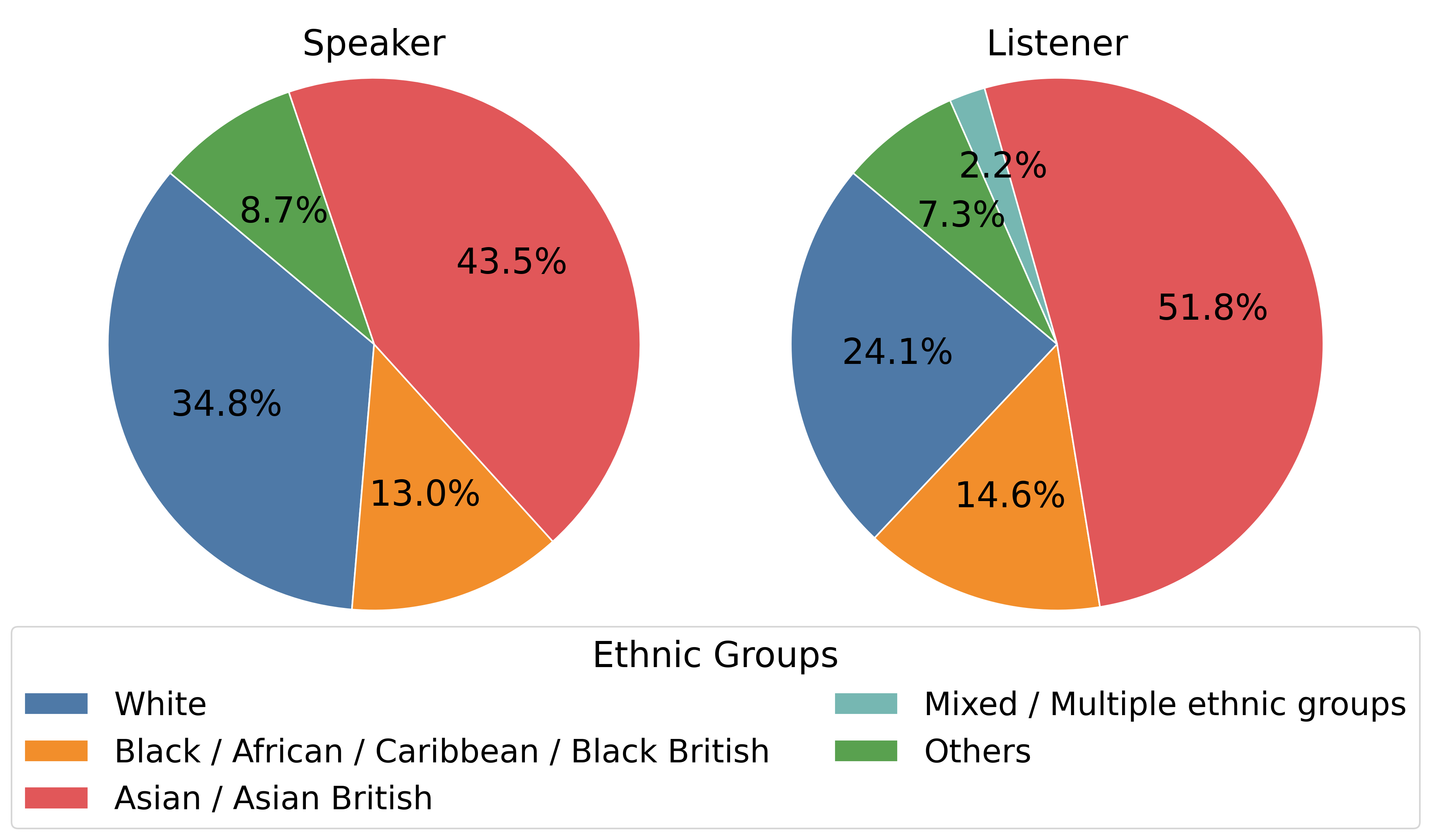}
    \caption{Statistics of participants' ethnic groups.}
    \label{fig:chart}
\end{figure}

\begin{table}[h]
    \centering
    \caption{Statistics of different interaction topics of the MARS dataset. `s' denotes seconds and `m' denotes minutes.}
    \begin{adjustbox}{width=0.86\columnwidth,center}
        \begin{tabular}{lccc}
            \toprule
             \multirow{2}*{Topics} & \multicolumn{3}{c}{Time} \\
            \cline{2-4}
            & Min & Max & Average\\
            \midrule
            Cultural differences & 195s & 530s & 339s \\
            Movie scene sharing & 143s & 573s & 321s \\
            Policy changes & 153s & 726s & 336s \\
            Quizzes and games & 191s & 415s & 311s \\
            Scenario-based interviews & 161s & 518s & 305s \\
            \midrule
            Overall interactions & 19m'50s & 40m'49s & 27m'16s \\
            \bottomrule
        \end{tabular}
    \end{adjustbox}
    \label{tab:statistics}
\end{table}

\section{Challenge Tasks}

\noindent Given a short-term behaviour $b(S_n)^{t_1,t_2}$ expressed by a speaker $S_n$ at the time period $[t_1,t_2]$, the REACT 2025 challenge follows the similar purpose and form as the REACT 2023 \cite{song2023react2023} and REACT 2024 \cite{song2024react} challenges, focusing on two Multiple Appropriate Facial Reaction Generation (MAFRG) tasks (e.g., offline Multiple Appropriate Facial Reaction Generation (offline MAFRG) and online Multiple Appropriate Facial Reaction Generation (online MAFRG)). The formal definition them is provided by \cite{song2023multiple}.

\textbf{Task 1: Offline Appropriate Facial Reaction Generation:} The offline MAFRG task aims to develop a machine learning model $\mathcal{H}$ that takes the entire speaker behaviour sequence $b(S_n)^{t_1,t_2}$ as the input, and generates multiple ($M$) realistic spatio-temporal appropriate facial reactions (AFRs) $p_f(L \vert b(S_n)^{t_1,t_2})_1, \cdots, p_f(L \vert b(S_n)^{t_1,t_2})_M$, where $p_f(L \vert b(S_n)^{t_1,t_2})_m$ is a multi-channel time-series consisting of AUs, facial expressions, valence and arousal states representing the $m_\text{th}$ predicted AFR. As a result, $M$ facial reactions are required to be generated for the task given each input speaker behaviour.

\textbf{Task 2: Online Appropriate Facial Reaction Generation:} The online MAFRG task aims to develop a machine learning model $\mathcal{H}$ that estimates each frame (i.e.,  $\gamma_\text{th} \in [t_1,t_2]$ frame) of the listener's facial reaction by only considering the $\gamma_\text{th}$ frame and its previous frames expressed by the corresponding speaker (i.e., ${t_1}_\text{th}$ to $\gamma_\text{th}$ frames in $b(S_n)^{t}$), rather than taking all ${t_1}_\text{th}$ to ${t_2}_\text{th}$ frames into consideration. The model is expected to gradually generate all facial reaction frames to form multiple ($M$) appropriate and realistic / naturalistic spatio-temporal facial reactions $p_f(L \vert b(S_n)^{t_1,t_2})_1, \cdots, p_f(L \vert b(S_n)^{t_1,t_2})_M$, where $p_f(L \vert b(S_n)^{t_1,t_2})_m$ is a multi-channel time-series consisting of AUs, facial expressions, valence and arousal state representing the $m_\text{th}$ predicted facial reaction. As a result, $M$ facial reactions are required to be generated for the task given each input speaker's behaviour. 











\section{Challenge Corpora}
\label{sec:dataset}

\textbf{Dataset.} The Multi-modal \textbf{M}ultiple \textbf{A}ppropriate \textbf{R}eaction in \textbf{S}ocial Dyads (MARS) Dataset is the first multi-modal dataset that are specifically collected for MAFRG tasks. It comprises 137 human-human dyadic interaction multi-modal (audio, visual and EEG) clips recorded from 23 speakers and 137 listeners (the details of these 155 participants are provided in Table \ref{tab:demographics} and Fig. \ref{fig:chart}). These involved participants can be categorized into two roles: speakers and listeners. Specifically, each clip capturing a pair of human speaker and listener's audio, face, and EEG behaviours in separate files, resulting in 270 multi-modal recordings (clips) whose durations range from 20 to 35 minutes (the statistics of the MARS dataset is provided in Table. \ref{tab:statistics}). Here, each multi-modal recording contains 23 distinct sessions covering five main topics conducted in a fixed order, including \textbf{cultural differences (four sessions per clip), movie scene sharing (four sessions per clip), policy changes (four sessions per clip), quizzes and games (eight sessions per clip), as well as scenario-based interviews (three sessions per clip)}. As a result, we split all clips into 2856 multi-modal session pairs (5712 sessions). During the recording of each clip, a pair of speaker and listener are situated in separate rooms to interact with each other on Microsoft Teams through screens. Prior to that, two volunteers assist both speaker and listener in setting up their cameras, microphones and EEG sensors (MUSE-2), while adjusting the webcam to optimally capture them within the recording frame. Then, the speaker is informed to start the conversation once the recordings of audio, video and EEG signal commence, where each speaker is responsible for initiating and directing the discussions with their corresponding listener, maintaining consistent semantic contexts through the pre-designed conversational topics (illustrated in Fig. \ref{fig:diagram}). Consequently, a set of diverse verbal and non-verbal behaviour reactions expressed by different listeners under a consistent context are considered appropriate in response to each speaker behaviour, which shapes the designed semantic context. Throughout the conversation session, the two volunteers monitor the interaction from a third room to ensure that the conversation is under the designed interaction control, and to immediately handle the unexpected interruptions (i.e., network interruption). At the end of the conversation, the volunteers back to the recording rooms to stop the recordings and assist participants in removing their wearable equipments. All clips of the MARS dataset were collected primarily from students and staff at the University of Leicester, United Kingdom, between June to October 2024. 

\textbf{Provided data:} Our MARS dataset includes the following \textbf{raw and pre-processed data}: (1) original videos; (2) cropped face videos that still includes expressions and head movements; (3) original audio clips; (5) textual transcripts; (6) EEG clips; and (7) the age, gender, race, and education level of each participant. In addition, we also provide \textbf{frame-level facial and audio descriptors} including 15 AUs extracted using \cite{song2022gratis,luo2022learning}, eight facial expression probabilities and valence/arousal intensities extracted using EmoNet \cite{toisoul2021estimation}, 58-dimensional 3D Morphable Model (3DMM) coefficients (i.e., 52 facial expression coefficients, three pose coefficients and three translation coefficients) extracted using  Faceverse \cite{wang2022faceverse} as well as 768-dimensional audio features extracted using wav2vec 2.0 \cite{baevski2020wav2vec}.

\textbf{Ground-truth labels.} During data recording, the semantic contexts are carefully controlled through the 23 distinct sessions, each of which is guided by a few pre-defined sentences posted by the speaker. This provides a consistent session-specific context across dyadic interactions between different speakers and listeners. More specifically, for each speaker behaviour expressed in a specific session, we define all facial reactions expressed by different listeners under the same session to be appropriate facial reactions (i.e., ground-truth) for responding to it. \textit{As a result, the length of different appropriate facial reactions (i.e., different listeners' facial behaviours expressed in a specific session) for responding to a speaker behaviour can be varied}. In summary, our MARS dataset includes the following labels: (1) ground-truth real appropriate facial reactions (GT real AFRs) for each speaker behaviour clip; and (2) self-reported personality traits based on the 10 question Big-Five inventory.



    


\textbf{Ethical consideration.} Ethical approval for our MARS dataset was obtained at the University of Leicester. Prior to data acquisition, participants are required to read and sign the consent form. Subsequently, they complete the personality questionnaire and are informed about the recording equipments and their use. Finally, all participants are compensated with Amazon vouchers for their participation in the study. During the challenge, all participating teams are required to sign an EULA to access the dataset.

\section{Evaluation Metrics}

\noindent In this challenge, the submitted models are expected to generate two types of outputs for representing each facial reaction: (i) 25 facial attribute time-series; and (ii) a 2D facial image sequence. We followed \cite{song2023multiple,song2023react2023} to comprehensively evaluate three aspects of the generated facial reaction attributes: (i) \textbf{Appropriateness} based on two metrics, \textbf{FRCorr}: Concordance Correlation Coefficient (CCC) and \textbf{FRDist}: Dynamic Time Warping (DTW); (ii) \textbf{Diversity}: \textbf{FRVar}, and \textbf{FRDiv}; and (iii) \textbf{Synchrony}: the Time Lagged Cross Correlation (TLCC), called \textbf{FRSyn} in this challenge. Also, the \textbf{Realism} of the generated facial reaction video clips is assessed using the Fréchet Inception Distance (FID), denoted as \textbf{FRRea}.

As detailed in Sec. \ref{sec:dataset}, the lengths of real AFRs expressed by different listeners for responding to a speaker behaviour can be varied. In this sense, we trained a Transformer-based variational encoder-decoder framework to summarise every arbitrary-length real AFRs into a fixed-length sequence of tokens, allowing to represent all variable-length real facial reactions with fixed-length token sequences in a latent space. This way, each generated AFR can be directly compared with all corresponding real AFRs of varied lengths in this latent space for computing \textbf{FRDist} and \textbf{FRCorr} metrics. Then, the transformer decoder can decode the token sequence as facial reaction image sequences of any required length.


\section{Baseline Models}


\noindent To enable fair and easy comparison, we first re-employ an open-source Trans-VAE designed for both online and offline MAFRG tasks in previous challenges \cite{song2023react2023,song2024react}. We also provide the generic MAFRG model of the state-of-the-art diffusion-based MAFRG approach (i.e., PerFRDiff \cite{zhu2024perfrdiff}) as both online and offline MAFRG baselines. In addition, we provide the REGNN  \cite{xu2023reversible} as the offline baseline.

\begin{figure}[tb!]
    \centering
    \includegraphics[width=0.96\columnwidth]{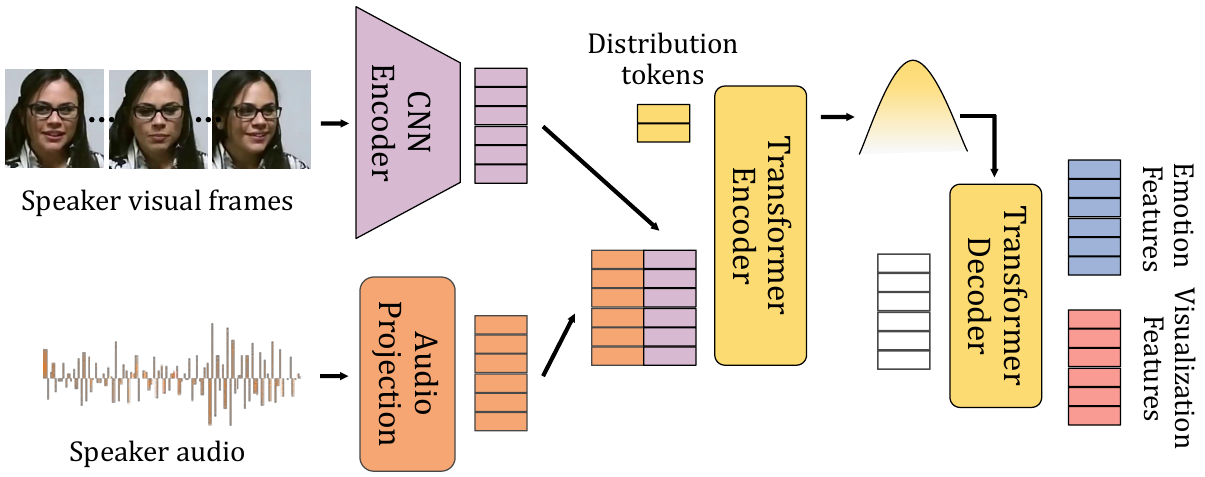}
    \caption{Overview of the Trans-VAE baseline.}
    \label{fig:Trans-VAE}
\end{figure}

\textbf{Trans-VAE \cite{song2023react2023,song2024react,luo2024reactface}:} We re-employ the same Trans-VAE baseline used in previous challenges \cite{song2023react2023,song2024react} to this challenge. This baseline is inspired by \cite{luo2024reactface}, which follows the similar architecture as the TEACH \cite{athanasiou2022teach}. As shown in Fig.~\ref{fig:Trans-VAE}, it is made up of (i) a \textbf{CNN encoder} extracting facial reaction-related features from the input speaker facial image sequence; (ii) a \textbf{Transformer encoder} combining the learned facial and baseline audio features (768-dimensional features) extracted from the speaker audio behaviours using wav2vec 2.0 \cite{baevski2020wav2vec}, based on which an Gaussian Distribution is learned to describe multiple predicted AFRs for responding to the input speaker behaviour; and (iii) a \textbf{Transformer decoder} that samples two types of facial reaction representations from the learned distribution: 1) a set of 3D Morphable Model (3DMM) coefficients \cite{wang2022faceverse};
and 2) an multi-channel facial attribute time-series. 
Please refer to \cite{song2023react2023} for more details

\begin{figure}[tb!]
    \centering
    \includegraphics[width=0.96\columnwidth]{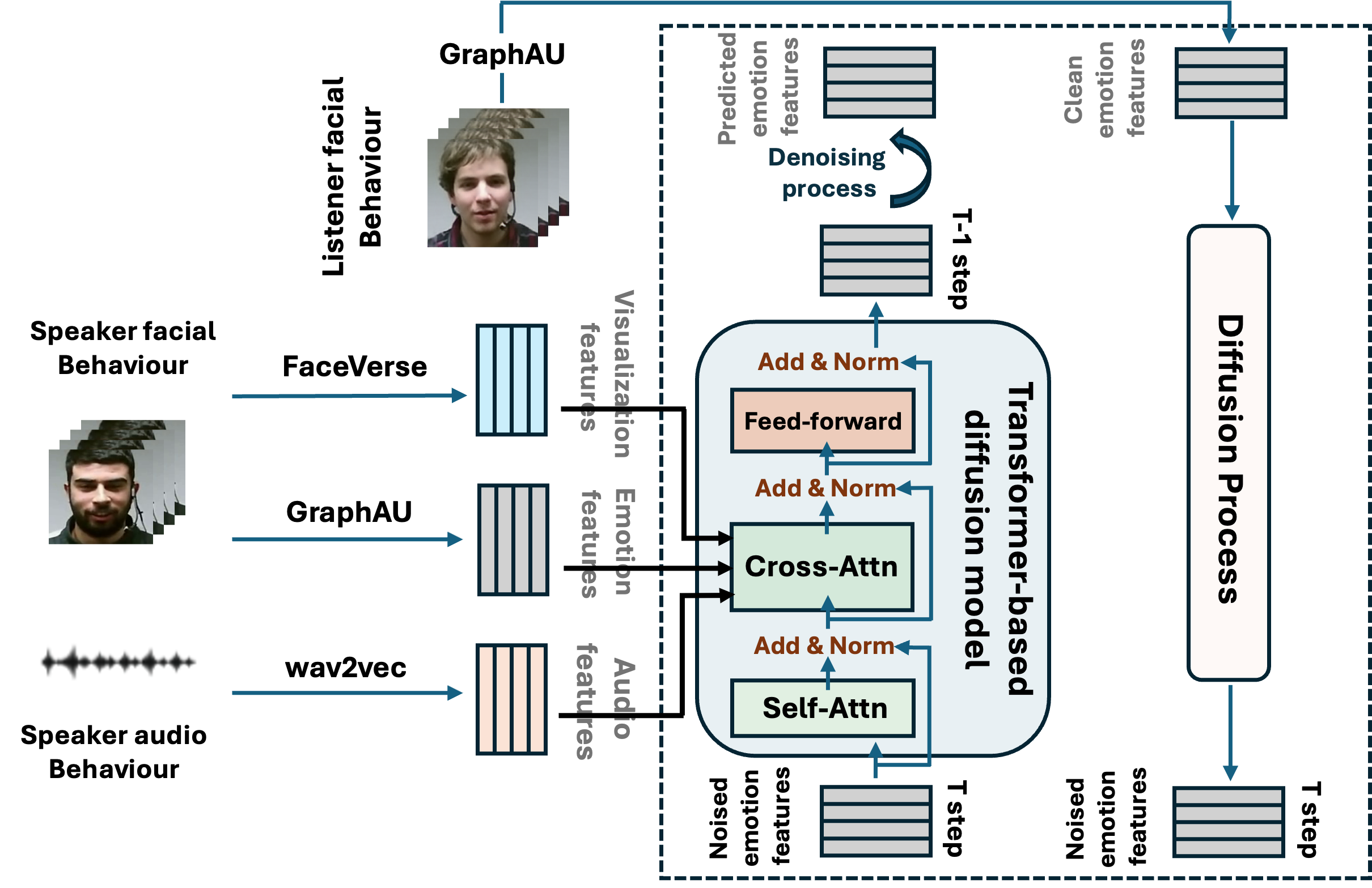}
    \caption{Overview of the PerFRDiff baseline.}
    \label{fig:perfrdiff}
\end{figure}

\begin{figure}[tb!]
    \centering
    \includegraphics[width=0.96\columnwidth]{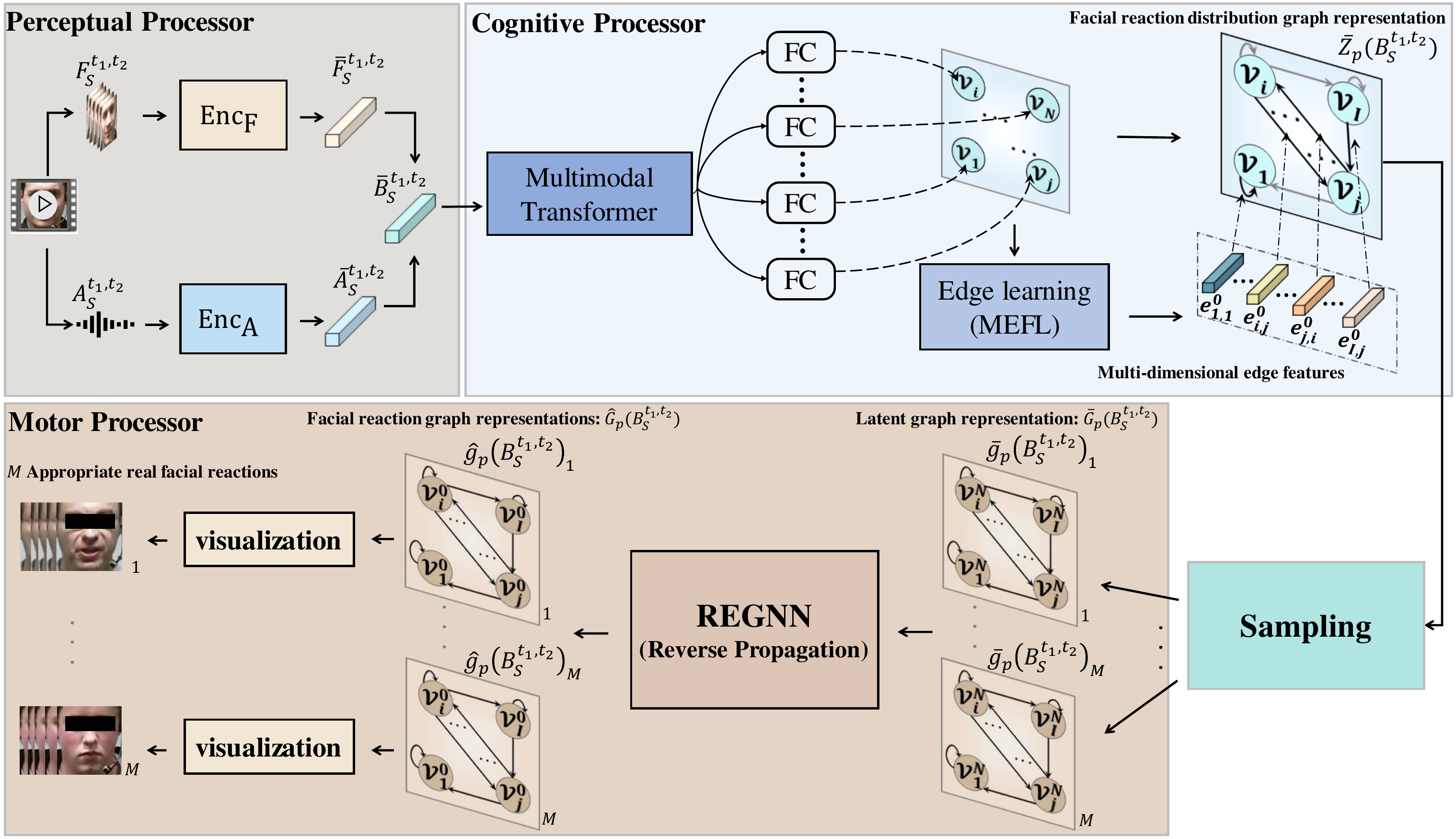}
    \caption{Overview of the REGNN baseline.}
    \label{fig:REGNN}
\end{figure}

\textbf{PerFRDiff \cite{zhu2024perfrdiff}:} We re-train the generic MAFRG model (i.e., a transformer-based conditional diffusion model) to separately conduct online task and offline MAFRG task. As demonstrated in Fig. \ref{fig:perfrdiff}, it employs three modality-specific linear projectors to individually map the input multi-modal speaker behaviours including audio features extracted from Wav2vec 2.0 \cite{baevski2020wav2vec}, raw multi-channel facial attribute time-series and raw 3DMM coefficients sequence, to a shared latent space, serving as conditional inputs to the diffusion Transformer. Then, the model follows the standard denoising diffusion process to gradually denoise clean AFRs, where cross-attention operations are used to process the conditions (i.e., speaker behaviour features). To accommodate variable-length speaker behavior sequences across batches, we apply standard padding masks in the attention layers to prevent information leakage from padded positions, following common practices in sequential modeling.

\textbf{Reversible Graph Neural Network (REGNN) \cite{xu2023reversible}:} We additionally employ the REGNN \cite{xu2023reversible} as the offline MAFRG baseline. As illustrated in Fig.~\ref{fig:REGNN}, it consists of three main modules: (i) a \textbf{Perceptual Processor} that encodes the input speaker audio-facial behaviour as a pair of latent audio and facial representations; (ii) a \textbf{Cognitive Processor ($\text{Cog}$}) that predicts a Gaussian Mixture Graph Distribution describing all AFRs in response to the input speaker behaviour; and (iii) a Reversible GNN-based \textbf{Motor Processor} that samples an appropriate facial reaction from the predicted appropriate facial reaction distribution. During training, the Reversible GNN encodes all real AFRs of each input speaker behaviour as a ground-truth real AFR distribution, enforcing the perceptual and cognitive processors to predict the same distribution from the input speaker behaviour. As a result, the \textit{one-to-many mapping} training problem is re-formulated as a \textit{one-to-one mapping} problem. 

\section{Baseline Results}

\begin{table}[t]
  \centering
  \caption{Baseline results achieved on the test set of MARS dataset, where B\_Random randomly samples $\alpha = 10$ facial reaction sequences from a Gaussian distribution; B\_Mime generates facial reactions by mimicking the corresponding speaker facial behaviours; B\_MeanSeq and B\_MeanFr generate facial reactions by average the sequence- and frame-wise facial reactions in the training set, respectively.}
  \label{tab:baseline-results}
    \begin{adjustbox}{width=1\columnwidth}
  \begin{tabular}{lrrrrrr}
    \toprule
    Method & \multicolumn{2}{c}{\textbf{Appropriateness}} & \multicolumn{2}{c}{\textbf{Diversity}} & \textbf{Realism} & \textbf{Synchrony} \\
    \cmidrule(lr){2-3} \cmidrule(lr){4-5} \cmidrule(lr){6-6} \cmidrule(lr){7-7}
     & FRCorr ($\uparrow$) & FRDist ($\downarrow$) & FRDiv ($\uparrow$) & FRVar ($\uparrow$) & FRRea ($\downarrow$) & FRSyn ($\downarrow$) \\
    \midrule
    GT & 10 & 0.00 & 0.1876 & 0.0669 & -- & 48.66 \\
    B\_Random & 0.03 & 474.68 & 0.3342 & 0.1671 & --  & 46.64 \\
    B\_Mime & 0.52 & 206.02 & 0.0000 & 0.0766 & --  & 43.70 \\
    B\_MeanFr & 0.00 & 205.65 & 0.0000 & 0.0000 & -- & 49.00 \\
    \midrule
    \multicolumn{7}{l}{\bfseries Offline Results} \\
    \midrule
    Trans-VAE & 0.30 & 181.72 & 0.0076 & 0.0083 & -- & 49.00 \\
    PerFRDiff & 0.58 & 217.32 & 0.1596 & 0.0862 & -- & 48.66 \\
    REGNN & 0.54 & 155.49 & 0.0012 & 0.0048 & -- & 44.81 \\
    \midrule
    \multicolumn{7}{l}{\bfseries Online Results} \\
    \midrule
    Trans-VAE & 0.25 & 212.22 & 0.1082 & 0.0564 & -- & 48.20 \\
    PerFRDiff & 0.55 & 222.69 & 0.1506 & 0.0841 & -- & 46.56 \\
    \bottomrule

  \end{tabular}
\end{adjustbox}
\end{table}

As shown in Table \ref{tab:baseline-results}, all baselines largely outperformed the B\_Random and B\_MeanFr, showing that they can to some extent generate meaningful AFRs from different speaker behaviours. Here, we found that the $B\_Mime$ setting achieved surprising FRCorr performance by simply generating the corresponding speaker facial behaviours, which can be explained by the fact that human listeners often exhibit facial behaviours that mirror those of the corresponding speakers, i.e., facial mimicry \cite{dimberg1982facial}.

\textbf{Offline MAFRG task:} As compared in the table, the generic MAFRG model of the PerFRDiff and REGNN achieved relatively decent performances over Trans-VAE. Specifically, the PerFRDiff achieved the highest FRCorr (0.58) with best diversity (more than 10 times for both metrics), indicating that it can generate multiple diverse facial reactions that show relatively similar trend as GT real AFRs. Meanwhile, the REGNN achieved similar FRCorr (0.54) but the lowest FRDist (155.49) and FRSyn, suggesting that the AFRs generated by it not only have similar trend but also closely aligned with the GT real AFRs.

\textbf{Online MAFRG task:} Similarly, the PerFRDiff also achieved clearly better performances over the Trans-VAE baseline in terms of all metrics except the slightly higher FRDist (222.69 vs. 212.22). Importantly, the PerFRDiff still achieved much higher FRCorr (0.55 vs. 0.25) with better facial reaction diversities and the synchrony between the input speaker behaviours and the generated facial reactions, validating its capability in simulating multiple appropriate human-style facial temporal patterns for responding conversational partners, while maintaining their diversity.

\section{Participation and Conclusion}

This paper introduces REACT 2025 Challenge in conjunction with the ACM Multimedia 2025 conference, which focuses on multiple appropriate facial reaction generation under various video conference-based dyadic interactions scenarios. 
Challenge participants were given access to the training and validation sets to develop their ML models, with a challenge guideline and a baseline paper released to provide more details. Each participating group were allowed to submit their models and results for the test set up to five times. The submitted models was automatically evaluated. Our evaluation protocol strictly ranked all participant models under the same settings by evaluating two aspects of the facial reactions generated by their models: appropriateness, diversity, realism and synchrony. We hope that both the new challenge dataset and baseline code, as well as the methodologies and results of the participating teams, will serve as a valuable stepping stone for researchers and practitioners interested in  generative AI and automatic facial reaction generation.

\bibliographystyle{ACM-Reference-Format}
\bibliography{bibliography}

\end{document}